\documentclass[10pt,twocolumn,letterpaper]{article}
\usepackage{graphicx}
\usepackage{amsmath}
\usepackage{amssymb}
\usepackage{booktabs}
\usepackage{amssymb}
\usepackage{balance}
\usepackage[pagenumbers]{cvpr}

\graphicspath{ {images/} }

\newcommand{\ssymbol}[1]{^{\@fnsymbol{#1}}}
\def\@fnsymbol#1{\ensuremath{\ifcase#1\or *\or \dagger\or \ddagger\or
   \mathsection\or \mathparagraph\or \|\or **\or \dagger\dagger
   \or \ddagger\ddagger \else\@ctrerr\fi}}

\usepackage[pagebackref,breaklinks,colorlinks]{hyperref}

\usepackage[capitalize]{cleveref}
\crefname{section}{Sec.}{Secs.}
\Crefname{section}{Section}{Sections}
\Crefname{table}{Table}{Tables}
\crefname{table}{Tab.}{Tabs.}

\def\cvprPaperID{5928} 
\def\confName{CVPR}
\def\confYear{2023}

\begin{document}

\title{MINTIME: Multi-Identity Size-Invariant Video Deepfake Detection}

\author{Davide Alessandro Coccomini$^1$, Giorgos Kordopatis Zilos$^2$, Giuseppe Amato$^1$, Roberto Caldelli$^{3,4}$, \\ Fabrizio Falchi$^1$, Symeon Papadopoulos$^2$, Claudio Gennaro$^1$ \\
$^1$ISTI-CNR \quad $^2$ITI-CERTH \quad $^3$CNIT \quad $^4$Universitas Mercatorum \\
{\tt\small \{davidealessandro.coccomini, giuseppe.amato, fabrizio.falchi, claudio.gennaro\}@isti.cnr.it} \\
{\tt\small \{georgekordopatis, papadop\}@iti.gr} \quad {\tt\small roberto.caldelli@unifi.it}}
\maketitle
\begin{abstract}
In this paper, we introduce MINTIME, a video deepfake detection approach that captures spatial and temporal anomalies and handles instances of multiple people in the same video and variations in face sizes. Previous approaches disregard such information either by using simple a-posteriori aggregation schemes, i.e., average or max operation, or using only one identity for the inference, i.e., the largest one. On the contrary, the proposed approach builds on a Spatio-Temporal TimeSformer combined with a Convolutional Neural Network backbone to capture spatio-temporal anomalies from the face sequences of multiple identities depicted in a video. This is achieved through an Identity-aware Attention mechanism that attends to each face sequence independently based on a masking operation and facilitates video-level aggregation. In addition, two novel embeddings are employed: (i) the Temporal Coherent Positional Embedding that encodes each face sequence's temporal information and (ii) the Size Embedding that encodes the size of the faces as a ratio to the video frame size. These extensions allow our system to adapt particularly well in the wild by learning how to aggregate information of multiple identities, which is usually disregarded by other methods in the literature. It achieves state-of-the-art results on the ForgeryNet dataset with an improvement of up to 14\% AUC in videos containing multiple people and demonstrates ample generalization capabilities in cross-forgery and cross-dataset settings. The code is publicly available at \url{https://github.com/davide-coccomini/MINTIME-Multi-Identity-size-iNvariant-TIMEsformer-for-Video-Deepfake-Detection}
\end{abstract}

\section{Introduction}
\label{sec:intro}


We are witnessing day by day an increasingly rapid evolution in the field of synthetic multimedia content generation. With the continuous advancement of Generative Adversarial Networks (GANs)\cite{NIPS2014_5ca3e9b1}  capable of generating highly credible images and videos, distinguish reality from fiction is a daunting task at best. Among synthetic content, the most dangerous are those involving human faces, i.e., deepfakes. Exploiting synthetic images and videos of people or manipulating existing ones has generated several defamation campaigns against public figures, celebrities or even regular citizens whose lives have been undermined by manipulated content\footnote{https://edition.cnn.com/interactive/2019/01/business/pentagons-race-against-deepfakes/}. 

\begin{figure}[t]
    \centering
    \includegraphics[width=1\linewidth]{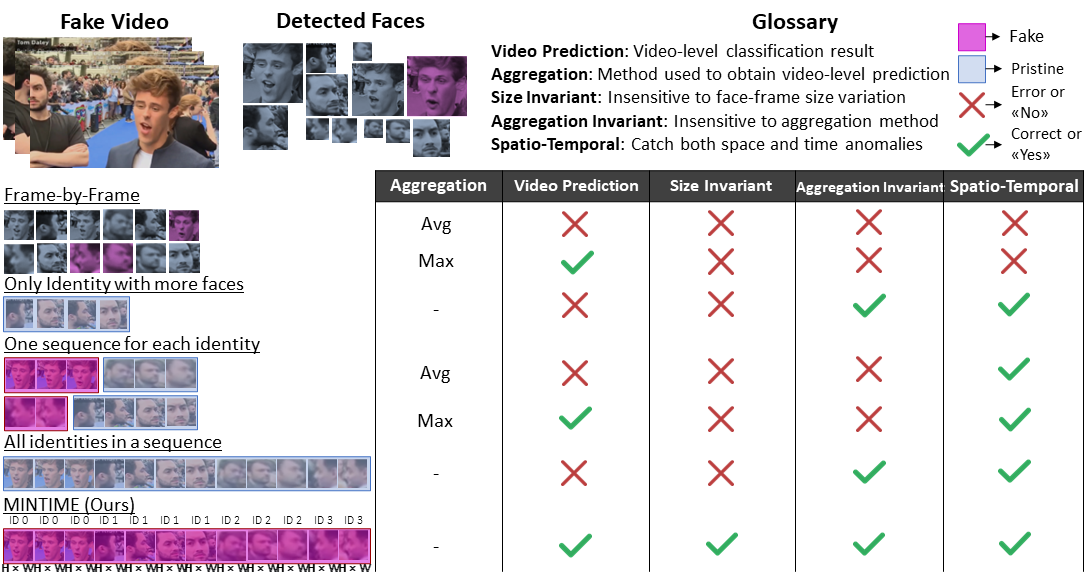} 
    \caption{The impact of the deepfake detection strategy in a case of video containing multiple identities and variations in the face-frame area ratio. The proposed approach is the only one capable of identifying spatio-temporal anomalies and at the same time effectively handling cases of multiple identities, variations in face size and without the use of aggregations that would impact the result.}
    \label{fig:comparison_deepfake}
\end{figure}


In response to this sudden and uncontrollable evolution, many efforts have been made to counteract deepfakes by implementing a multitude of deepfake detection systems 
based on a variety of approaches. 
However, it is evident that being able to distinguish manipulated from pristine content introduces many challenges and there are still many open issues. 
In this work, we focus on those issues that we consider crucial for detecting deepfakes `in the wild' and have yet to attract the research community's attention.
Previous studies \cite{ftcn, realforensics,meverdeepfake} have pointed that identifying temporal anomalies and inconsistencies between two frames of the same video, is fundamental for successful deepfake detection. Much work, however, tends to focus on spatial anomalies only, handling videos with a frame-by-frame classification approach and relying on naive aggregation schemes to extract video-level predictions\cite{mesonet,Bayar2016ADL,10.1145/3549555.3549588, 10.1145/3512732.3533582}.
When videos are analyzed frame-by-frame or divided into separately classified sequences, the problem of their aggregation to obtain a video-level prediction also arises. We can divide the approaches into a-posteriori aggregation, i.e. those that use static functions as average or maximum to obtain the final result, and internal aggregation, i.e. those that let the model analyze the entire video as it is and so decide internally the final video-level prediction. Methods based on the first approach are very sensitive to the choice of aggregation function as shown in \cite{preprocessing, meverdeepfake}.

An additional problem resides in a vulnerability that attackers can exploit to deceive a deepfake detector. In cases of videos with multiple distinct people (identities) appearing together \cite{Le2021OpenForensicsLC}, an attacker could decide to manipulate only one of them. However, if the detection is carried out \emph{en bloc} for all the detected faces, the negative contribution to the final prediction made by the fake faces could be `overshadowed' by the non-manipulated ones, thus deceiving the system. One possible approach would be to split the video into several clips, at least one for each identity, but this would require multiple forward passes of the model, and there are currently no techniques that can handle any number of identities in a single one. These approaches also rely on an aggregation policy for the distinct predictions to obtain an overall video-level prediction, which significantly affects the final result. Furthermore, in general, previous works in the literature have put minimal effort into this specific situation, which, however, is extremely frequent in real-world verification tasks.

Typically in deepfake detection systems, faces are detected from the video or image to be classified, and before being given as input to a classification model, it is resized to be uniform with all the others. This may result in an essential loss of information, namely the size of the subject's face as a ratio to the rest of the scene. Such information could be exploited to build models robust to environmental clutter, e.g., a sudden change of distance from the camera or a blur resulting from an original small image, which can distinguish better anomalies introduced by the deepfake generation process.

Finally, the creation of a deepfake also tends to introduce specific anomalies within images and videos. Hence, deepfake detectors often end up learning to recognize only the ones included in the training dataset and are therefore ineffective  when dealing with unseen manipulations, demonstrating poor generalization capability \cite{9156378, 10.1007/978-3-030-58610-2_6, 10.1007/978-3-030-58574-7_7, Haliassos2021LipsDL, Li2020FaceXF, fornow, 10.1145/3512732.3533582}. Furthermore, many approaches may be ineffective in real-world cases because they are trained and validated on very constrained situations where, for example, there is only a single subject in the video or people tend to always stay at the same distance from the camera \cite{wilddeepfake}. 


To overcome the above challenges, we present the Multi-Identity size-iNvariant TIMEsformer (MINTIME) for video deepfake detection. The main novelties presented by our approach, which are also illustrated in Figure \ref{fig:comparison_deepfake}, are:
\begin{itemize}
    \item Ability to identify both spatial and temporal anomalies through a combination of TimeSformer\cite{Bertasius2021IsSA} and Convolutional Neural Networks (CNNs), unlike other hybrid models designed for deepfake detection\cite{10.1145/3549555.3549588,10.1007/978-3-031-06433-3_19} working exclusively in a frame-by-frame manner.
    \item Ability to handle multiple people in the same video through the introduction of a novel positional embedding technique, namely Temporal Coherent Positional Embedding inspired by the traditional one but capable of maintaining both spatial and temporal coherence and a new type of attention, namely Identity-aware attention, capable of keeping track of the identity to which each face detected in the video refers.
    \item Ability to handle variations in the face-frame area ratio through the introduction of size embeddings that keep track of the ratio between the detected face area and the entire frame at each instant of time.
    \item Being unaffected by aggregation strategies through internal aggregation obtained by analyzing the entire video in a single sequence and let it decide the video-level prediction and being able to handle multi-identity videos even with a single forward pass. This way, the model directly returns a single prediction for the entire video without requiring additional post-processing.
\end{itemize}
The performance of the proposed system has been evaluated in a multitude of different contexts, with an improvement of up to 14\% AUC in videos containing multiple identities. It has been also validated in some cross-forgery and cross-dataset scenarios, and exceeds the state-of-the-art in all contexts, with improvements in the AUC score of up to 22\% in some cases, also demonstrating a high level of generalization.

\section{Related Work}
The growing interest in deepfakes has fuelled the emergence of numerous solutions to detect them in a variety of ways\cite{emergence}. Generally speaking, what practically all these methods seek to achieve is to correctly classify a video as pristine or deepfake by looking for anomalies of various kinds in it. 
As we focus on video deepfake detection, we summarize previous works into two categories based on the type of anomalies they focus on.

\textbf{Space-Only}
These methods often treat the video as a series of frames by performing a separate classification for each of them and then aggregating them into a final overall classification using an a-posteriori aggregation scheme such as the average or maximum function. 
These approaches focus primarily on identifying specific anomalies introduced by deepfake generation methods, namely those of a spatial nature. Even though they are most suited for image deepfake detection, where there is no need for detecting temporal anomalies, they are often applied to videos in a frame-by-frame manner. Most of them are based on well-established Deep Learning techniques such as CNNs\cite{faceforensics, Bayar2016ADL, mesonet, 8644886} but recently also some attention-based approaches have been proposed \cite{9156378, Wang2021RepresentativeFM, multiattentional}. Some previous works also propose hybrid architectures that combine Vision Transformers with various types of CNNs as in \cite{10.1007/978-3-031-06433-3_19} exploiting an EfficientNet-B0\cite{efficientnet} as a features extractor to feed a Cross Vision Transformer\cite{crossvit}. Aggregation in this case is done by making the maximum among the predictions obtained on the individual frames. A similar approach but using an XceptionNet\cite{xception} is the proposal in \cite{10.1145/3549555.3549588}. However, all these methods are unable to pick up temporal anomalies that can be crucial in deepfake detection.

\begin{figure*}[t]
    \centering
    \includegraphics[width=1\linewidth]{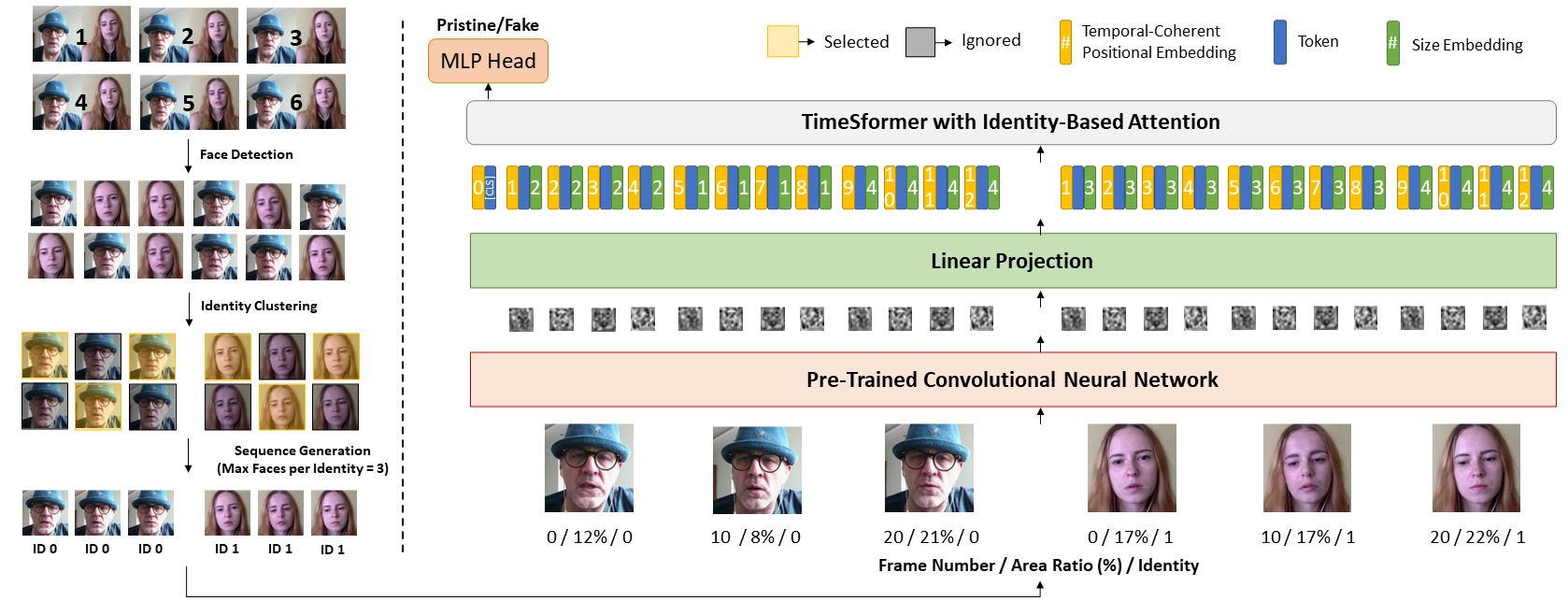}
    \caption{MINTIME overview: The preprocessing pipeline (left) starts with the detection of faces in the video, follows with the clustering of identities and then creates the input sequence. The sequence of faces is converted into features by the convolutional backbone (right), which once converted into tokens and concatenated to the embeddings, pass into the TimeSformer and finally into the MLP Head for the final classification.}
    \label{fig:mintime}
\end{figure*}

\textbf{Spatio-Temporal}
These approaches aim to 
capture temporal anomalies and obtain a video-level classification without any kind of aggregation. 
Some of them focus on specific types of spatio-temporal artifacts common in the case of deepfake videos, such as anomalous lip movement \cite{Haliassos2021LipsDL} or inconsistent eye-blinking \cite{8630787}, but they are clearly limited since they do not look for additional anomalies that may be present in other parts of the face. Proposals have also been made that exploit the optical flow of video \cite{9022558} or analyze the relationships between audio peaks and video content \cite{10.1145/3394171.3413570}. 
A more complete proposal was made by \cite{realforensics}, which uses a self-supervised approach to learn effective frame representations to train an audiovisual, cross-modal, student-teacher framework used then to train a deepfake detector. The authors of \cite{10.1145/3369412.3395070} also proposed a method that tries to detect inter-frame anomalies via LSTM networks looking for temporal inconsistencies. Finally, one of the most relevant methods focusing on temporal incoherence within videos is FTCN\cite{ftcn}, which uses a Temporal Transformer Encoder at its core. 
Although these methods efficiently handle videos by capturing spatio-temporal anomalies, they do not consider several important nuances of the problem. 
For example, they often construct the input sequence by selecting a single person from the video, even when there are more than one, and do not take into account the different face-frame area ratios that may occur or vary in them. An attempt to handle cases of multiple people in the video comes from \cite{10.1007/978-3-031-06433-3_19} in which identities are classified separately and if even one of them is manipulated then the whole video is considered fake but this method is frame-based and relies on an ad hoc aggregation scheme (average or max) also requiring a forward pass for each frame. Incorrect handling of these situations can lead to completely ignoring a manipulated subject in favor of a perfectly pristine one or mistaking variations in the face size for anomalies.
An aspect that has been partially exploited in deepfake detection is the identity of the subjects filmed in videos. Some works attempt to detect deepfake by identifying a person from temporal facial features, specific to how a person moves while talking \cite{9710044} or by the usage of  biometrics analysis techniques \cite{Agarwal2020DetectingDV}. 
However, these latter approaches remain particularly effective mainly in the case of a major manipulation of the person's identity and have limited capability in detecting videos manipulated by other types of deepfake generation methods. They are also often very specific to an identity and are ill-suited in cases where the subject under consideration is not a famous person and there are few videos depicting him or her.
\begin{figure}[t]
  \centering
   \includegraphics[width=1\linewidth]{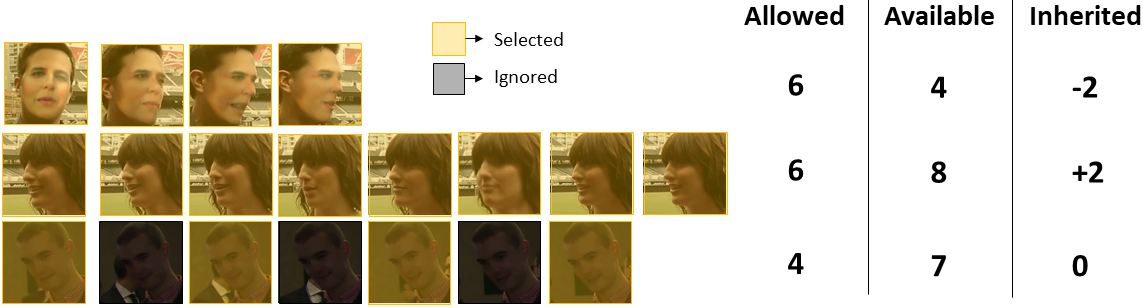}
   \caption{Sequence sampling when the number of allowed faces in the sequence is inherited through identities. The numbers on the right indicate, for each identity, how many faces can be used, how many are available, and how many have been inherited from or to another identity.}
    \label{figure:sequence_generation}
\end{figure}
\begin{figure*}[t]
    \centering
    \includegraphics[width=0.8\linewidth]{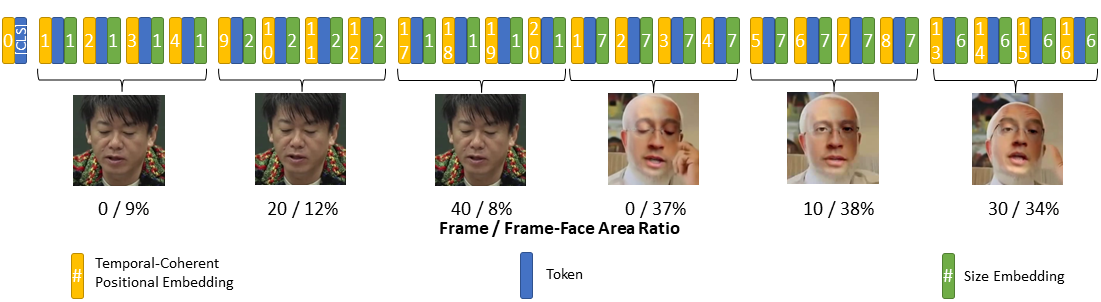}
    \caption{Overview of Temporal Coherent Positional Embedding and Size Embedding on a two-identities video frames.}
    \label{fig:embeddings}
\end{figure*}

\section{Proposed Approach}
The proposed Multi-Identity size-iNvariant TIMEsformer (MINTIME) architecture is illustrated in Figure \ref{fig:mintime}. It is capable of receiving as input a sequence of $N$ faces containing one or more identities and detect whether the video has been manipulated. MINTIME is versatile and able to efficiently adapt to a multitude of real world complex settings as described in Section \ref{sec:intro}.
In the following paragraphs we present in detail the proposed MINTIME novelties. 

\subsection{Adaptive Input Sequence Assignment}
\label{sec:sequence_assignment}
To enable the model to handle multiple identities within one video we generate a single input sequence composed by faces of all the considered identities. The identities are reordered according to the size of the faces within them, and the most important identities are given a higher number of slots to be exploited in the input sequence, in order to give more importance to faces that cover a larger area and are therefore likely to be more relevant in the video, as opposed to smaller faces. The length of the input sequence, and so the available slots, is fixed in our case to 16.
The number of faces for each identity is fixed and based on the number of identities in the video. If there is only one identity in the video, the faces detected by it will be used to fill the entire sequence, otherwise the sequence is filled by distributing the space between the various identities and giving higher priority to those containing faces that occupy a larger area. In case where there are fewer faces in the considered identity than necessary and no other identities are available, more empty images are added and then a mask is used to drive the calculation of attention properly, ignoring these blank images. In the case of longer sequences, however, uniform sampling is performed as shown in Figure \ref{figure:sequence_generation}. 
As a kind of input sequence diversification, this uniform sampling is performed in order to alternate various combinations of detected faces. Indeed, in this way at different epochs, the faces chosen for a given video are not necessarily always the same and the probability of catching the anomalies  increases. 
Formally, the output of this process will be the sets of tuples $\mathcal{X} = \{(\textbf{x}_1, I_1, s_1, t_1), (\textbf{x}_2, I_2, s_2, t_2), ..., $ $(\textbf{x}_N, I_N, s_N, t_N)\}$, where $N$ is the length of the input sequence. Each tuple contains the face image $\textbf{x}_i \in \mathbb{R}^{3 \times H \times W}$, where 3 is the number of channels of RGB image and $H$ and $W$ are its height and width, respectively, their corresponding identity depicted in the input video $I_i\in \mathbb{N}$, their face-frame area ratio $s_i\in (0., 1.]$ and their timestamps $t_i\in \mathbb{N}$ corresponding to the frame index. This set represents the information given as input to the system to perform the detection. 

\subsection{Backbone network}
We use a CNN backbone to generate features as input to the TimeSformer to help it more easily grasp the spatial information of the analyzed faces during training. In our case, an EfficientNet-B0 inspired by \cite{10.1007/978-3-031-06433-3_19} and an XceptionNet as in \cite{10.1145/3549555.3549588} were chosen as CNN backbones. The input images are then transformed into features by this CNN backbone, which
can be described as $C:\mathbb{R}^{3 \times H \times W} \rightarrow \mathbb{R}^{D \times H' \times W'}$. The resulting features is then $\textbf{z}_i = C(\textbf{x}_i) \in \mathbb{R}^{D \times H' \times W'}$, where $H'$ and $W'$ are the dimensions of the feature map resulting from the transformation and depend on the chosen backbone.

\subsection{Temporal Coherent Positional Embedding}
We evolved the positional embedding presented in the original Transformers paper\cite{attentionisallyouneed} in order to ensure temporal consistency between frames as well as spatial consistency between tokens. Tokens are numbered in such a way that two faces, of different identities but belonging to the same frame, have the same numbering. Temporal coherence is maintained locally by having an increasing numbering sequence related to the frames from which the faces have been detected. It is also maintained globally by being generated on the basis of the global distribution of frames of all identities in the video. In this way, a more significant variation simply due to a greater sparsity of the frames considered is not interpreted as an anomaly. We called this approach Temporal Coherent Positional Embedding (TCPE).
Figure \ref{fig:embeddings} depicts an example case where, although the two identities have the same number of faces, they are extracted from different frames and this information is encoded in this embedding. The intervals of integer values $\mathcal{E}_i \subset \mathcal{E} \subset \mathbb{N}$ to be assigned to the tokens related to the faces detected at frame $i$ is obtained as: 
$$ \mathcal{E}_i = \{i  T + 1, (i+1) T\} \quad \forall i \in \{0,\ldots, N-1\} $$
where $i$ is the frame index associated with the tokens derived from the face that needs to be enumerated, $N$ is the total number of faces detected and $T$ is the number of tokens for each face in the input sequence.
The numbering associated with tokens thus acquires both spatial meaning, i.e. incremental for tokens of the same face, and temporal meaning, i.e. incremental with respect to all faces of the video. This concept may be formally described by the function $TE(\cdot)$ that $TE:\mathbb{N}\rightarrow\mathbb{R}^D,\quad \textbf{z}_i'=\textbf{z}_i+TE(t_i)$.

\subsection{Size Embedding}
As explained in Section \ref{sec:intro}, not considering the variation of the face-frame area ratio can lead to misclassification errors. To handle this situation, we further enriched each token with a Size Embedding, a numeric index generated on the basis of the face-frame area ratio. The numerical value that the size embedding assigns to tokens belonging to a face is chosen by considering 20 intervals each with a width of $5\%$, in the range $[0\%, 100\%]$.
The set of all the intervals $\mathcal{S} \subset \mathbb{R}$ corresponds to the union of the $L$ subset obtained as:
$$ \mathcal{S}_i = [5i,\ 5(i+1)) \quad \forall i \in \{0,\ldots, L-1\}, \quad \mathcal{S}_i \subset \mathcal{S}$$
where $L$ is fixed to 20. 
For example, if a face has an area covering the 16\% of the entire frame, the size embedding associated with its tokens will be $3$ because it falls in the fourth interval $[15\%, 20\%)$. 
Formally, this is a function $SE(\cdot)$ that $SE:[0, 1]\rightarrow\mathbb{R}^D, \quad \textbf{z}_i'=\textbf{z}_i+SE(s_i)$.
\begin{figure}[t]
    \centering
    \includegraphics[width=0.7\linewidth]{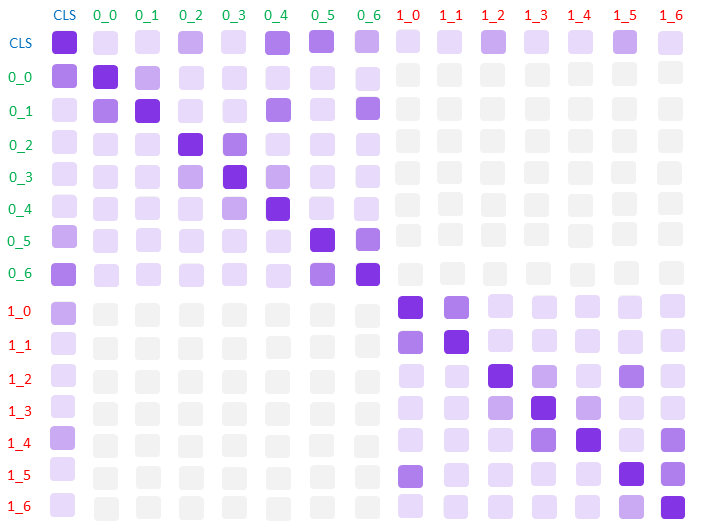}
    \caption{Identity-aware Attention on tokens for two-identities video. The naming $X\_Y$ stands for the identity ID and token ID respectively.}
    \label{fig:attention}
\end{figure}

\subsection{Identity-aware Attention}
We use the Divided Space-Time Attention TimeSformer, which was found to be the best performing one \cite{Bertasius2021IsSA}. Attention is calculated spatially between all patches in the same frame, but then it is also calculated between the corresponding patches in the next and previous frames. As far as spatial attention is concerned, no further effort is required for this to be applied to our case. Not being interested in capturing the relationships between faces of different identities, the calculation of temporal attention in our context is carried out exclusively between faces belonging to the same identity.
All faces, however, influence the CLS which is global and unique for all identities. 
Figure \ref{fig:attention} shows how attention is calculated exclusively by tokens referring to identity 0 faces (green), ignoring those referring to identity 1 faces (red) and vice versa, while all refer to the global CLS.

Formally, let $\textbf{Z} \in \mathbb{R}^{(NHW+1) \times D}$ be the set of all the tokens in which $\textbf{Z}^{I_0} \in \mathbb{R}^{1 \times D}$ is the global CLS token and $\textbf{Z}^{I_i} \in \mathbb{R}^{N^{I_i}HW \times D}$ is the set of all tokens related to identity $I_i$, where $N^{I_i}$ is the number of faces for identity $I_i$. Consider $M^{I_i} \in \{0,1\}^{(NHW+1) \times 1}$ as the mask of identity $I_i$ indicating to the model which faces attend to an identity and which to another during the attention calculation.
Then, our proposed Identity Attention (IA) can be formulated as: 
\begin{equation}\label{eq:identityattention}
    IA(\textbf{Z}^{I_i}, \textbf{Z}) =
    \begin{cases}
        Att(\textbf{Z}^{I_0}, \textbf{Z})  &  i=0 \\
        Att(\textbf{Z}^{I_i}, \textbf{Z} \odot M^{I_i})  &  i\neq0 \\
    \end{cases}
\end{equation}
where the attention calculation $Att(\cdot, \cdot)$
takes place according to \cite{Bertasius2021IsSA}, 
with the first argument corresponding to queries and the second to key-values, and $\odot$ denotes element-wise multiplication with broadcasting. 

\section{Evaluation}
\subsection{Dataset}

A study of the deepfake detection video datasets in the literature was first conducted to find one that would include temporal anomalies, various face-frame area ratios, videos with several people in the same scene at the same time, and varied in terms of deepfake generation methods and perturbations.  
For these reasons, ForgeryNet\cite{forgerynet} was chosen as the most suitable dataset for our experiments. It consists of 221,247 videos, 99,630 pristine, and 121,617 manipulated videos with a frame rate between 20 and 30 fps and variable duration. Of those, 11,785 videos include multiple identities with a maximum of 23 faces per video. We analyzed in the supplementary material, the face-frame area ratio of videos in this dataset discovering that it is very variable with videos containing faces covering an area up to almost, in some rare cases, even 100\% of the entire frame.

\subsection{Preprocessing}
All the faces of the depicted subjects are first detected, using the MTCNN face detector \cite{mtcnn}. The detected faces are then clustered by calculating the similarity between the face embeddings extracted via an InceptionResnetV1 pretrained on FaceNet\cite{facenet} and grouping them into identities, as proposed in \cite{preprocessing}. These identities are then used for the construction of the input sequence as illustrated in Section \ref{sec:sequence_assignment}. In order to ensure greater generalization capability, in the training phase, the sequences undergo a process of data augmentation in which many random perturbations inspired by \cite{forgerynet} are applied. We report more details about the preprocessing phase in the supplementary material.

\subsection{Training Setup}
MINTIME was trained in two main versions, a) MINTIME-EF, which uses an EfficientNet-B0\cite{efficientnet} as feature extractor and trained on DFDC dataset\cite{DFDC} as in \cite{10.1007/978-3-031-06433-3_19}, and b) MINTIME-XC, which uses an XceptionNet\cite{xception} as feature extractor (inspired by \cite{10.1145/3549555.3549588}) and trained on ForgeryNet images as in \cite{forgerynet}.
The two versions differ also in the training modality in fact the first was trained keeping fixed the whole convolutional backbone excluding the last 2 blocks, with a batch size of 8 and on an NVIDIA RTX 3060, while MINTIME-XC was trained End2End with a batch size of 32 in parallel on four NVIDIA A100 for 30 epochs.
The optimizer used is SGD with a learning rate of 0.01, which decays to 0.0001 using a cosine scheduler. The weight decay was set to 0.0001 as in \cite{Bertasius2021IsSA}.
All models were trained considering a maximum of two identities per input sequence, which does not limit the possibility of using more or fewer identities at inference time. The maximum number of faces we put in the input sequence was set to 16. Since the SlowFast\cite{Feichtenhofer_2019_ICCV} model trained in \cite{forgerynet} was not available, we retrained it starting from the model pretrained on Kinetics 400\cite{kinetics} in order to see how it behaved in certain contexts not reported in the original paper. This is the only training conducted in which a single identity per video is considered in order to emulate what has been done in \cite{forgerynet}.

\subsection{Metrics}
To evaluate the performance of our model, we used the accuracy and AUC metrics in most contexts. This is because they are widely used in the literature by previous methods and because they are highly indicative in a binary classification context such as this. Two additional metrics were also occasionally used in the tables below, namely the False Positive Rate (FPR) and the Maximum Accuracy Variation (MAV), calculated as $FPR = \frac{FP}{TN+FP}$ and $MAV = max(\mathcal{A}) - min(\mathcal{A})$, where FP and TN stand for False Positives and True Negatives respectively and $\mathcal{A}$ is the set of accuracy scores obtained by the model on all the classes. The FPR was chosen as a metric because it is important to have a system in the real-world that does not arise too many false detections while the MAV is used to give an idea of how much the models generalize among the various forgery methods.

\begin{table}
    \centering
    
    \resizebox{0.8\columnwidth}{!}{%
    \begin{tabular}{llll}
    \hline
        Model & Identities & Accuracy & AUC \\ 
        \hline
        SlowFast R-50$\ssymbol{2}$ \cite{Feichtenhofer_2019_ICCV}  & 1 & 82.59 & 90.86 \\
        SlowFast R-50$\ssymbol{3}$ \cite{Feichtenhofer_2019_ICCV}  & 1 & \textbf{88.78} & 93.88 \\ 
        X3D-M$\ssymbol{3}$ \cite{9156381} & 1 & 87.93 & 93.75 \\ 
        MINTIME-EF & 1 & 81.92 & 90.13 \\ 
        MINTIME-EF & 2 & 82.28 & 90.45 \\ 
        MINTIME-EF & 3 & 82.05 & 90.28 \\ 
        MINTIME-XC & 1 & 85.96 & 93.20 \\ 
        MINTIME-XC & 2 & 87.64 & \textbf{94.25} \\ 
        MINTIME-XC & 3 & 86.98 & 94.10 \\ 
    \hline
    \end{tabular}
    }
    \caption{Video-Level Evaluation on ForgeryNet Validation Set. The identities column
represents the number of considered identities during the inference. $\ssymbol{2}$ Indicate that the model has been trained in our setup. $\ssymbol{3}$ Indicate that the result is taken from \cite{forgerynet}.}
    \label{tab:forgerynet_validation}
\end{table}
\begin{table}
    \centering
    
    \resizebox{0.7\columnwidth}{!}{%
    \begin{tabular}{llll}
    \hline
        Model & Accuracy & AUC \\ \hline
        SlowFast R-50~\cite{Feichtenhofer_2019_ICCV} & 72.63 & 80.92 \\
        MINTIME-EF & 81.21 & 89.56\\ 
        MINTIME-XC & \textbf{86.68} & \textbf{94.12} \\ 
        \hline
    \end{tabular}
    }
    \caption{Evaluation on multi-identity only videos of ForgeryNet Validation Set. The models are all trained in our setup.}
    \label{tab:multi_identity}
\end{table}
\begin{table*}
    \centering
    \begin{tabular}{llllllllllll}
    \hline
        Model & \multicolumn{9}{c}{Forgery Method} & FPR & MAV \\
        ~ & Pristines & 1 & 2 & 3 &  4 & 5 & 6 & 7 & 8 & ~\\ \hline
        SlowFast R-50~\cite{Feichtenhofer_2019_ICCV} & 84.65 & 69.70 & 71.71 & 81.19 & 81.35 & \textbf{78.67} & \textbf{88.43} & 88.96 & \textbf{92.05} & 15.34 & 22.36\\
        MINTIME-EF & 85.84 & 70.05 & 69.75 & 74.55 & 82.05 & 78.14 & 79.59 & 91.49 & 77.03 & 14.16 & 21.74 \\ 
        MINTIME-XC & \textbf{88.15} & \textbf{79.94} & \textbf{84.64} & \textbf{82.17} & \textbf{84.05} & 77.59 & 85.37 & \textbf{92.03} & 79.91 & \textbf{14.06} & \textbf{12.12} \\ \hline
    \end{tabular}
    \caption{Video-Level evaluation on ForgeryNet Validation Set. The models are all trained in our setup.}
    \label{tab:forgerynet_methods}
\end{table*}
\begin{table}
    \centering
    \resizebox{\columnwidth}{!}{%
    \begin{tabular}{llllllll}
    \hline
        ~ & ~ & \multicolumn{2}{c}{ID-replaced} & \multicolumn{2}{c}{ID-remained}  & ~ \\ 
        ~ & ~ & Accuracy & AUC & Accuracy & AUC & ~ & ~ \\ \hline
        X3D-M\cite{9156381} & ID-replaced & 87.92 & 92.91 & 55.25 & 65.59 & ~ \\ 
        ~ & ID-remained & 55.93 & 62.87 & 88.85 & 95.40 & ~ & ~ \\ \hline
        SlowFast R-50\cite{Feichtenhofer_2019_ICCV} & ID-replaced & \textbf{88.26} & 92.88 & 52.64 & 64.83  & ~ \\ 
        ~ & ID-remained & 52.70 & 61.50 & 87.96 & 95.47 & ~ & ~ \\ \hline
        MINTIME-EF & ID-replaced & 80.18 & 83.86 & 79.03 & 86.98  & ~ \\ 
        ~ & ID-remained & 63.13 & 66.26 & 89.22 & 95.02 & ~ & ~ \\ \hline
        MINTIME-XC & ID-replaced & 86.58 & \textbf{93.66} & \textbf{84.02} & \textbf{88.43}  & ~ \\
        ~ & ID-remained & \textbf{64.01} & \textbf{68.53} & \textbf{92.08} & \textbf{97.26} & ~ & ~ \\ \hline
    \end{tabular}
    } 
    \caption{Cross-Forgery Evaluation on ForgeryNet Validation Set. X3D-M and SlowFast R-50 results are taken from \cite{forgerynet}.}
    \label{tab:cross_forgery}
\end{table}

\section{Experiments}
\subsection{Comparison with State of the Art}
\paragraph{ForgeryNet Evaluation}
According to Table \ref{tab:forgerynet_validation}, MINTIME-XC outperforms the state-of-the-art on the ForgeryNet validation set in terms of AUC and is quite similar to what was obtained with a SlowFast in terms of accuracy, which is, however, limited to considering a single identity in the classification phase.  
All MINTIME models also prove to be particularly robust at analyzing videos considering a variable number of identities without a significant loss of overall accuracy.

By testing the trained models considering only videos containing more than one person in the scene, MINTIME-XC significantly outperforms the trained state-of-the-art models by correctly classifying most of the videos considered as shown in Table \ref{tab:multi_identity}.

Interestingly, SlowFast R-50 trained considering only one identity per video performs poorly on multi-identity videos as it is heavily influenced by the choice of this single identity to be analyzed. This is certainly one of the most interesting results obtained as it concretely highlights how necessary it is to create models capable of handling multiple identities in the same video and not simply rely on selection criteria for these. In our case, we consider identities containing faces that cover larger areas of the video to be more important but we report a deeper analysis of the impact of this choice in the supplementary material where we show that our approach has no loss of performance when this choice is varied. We also report our experiments highlighting the impact of Temporal Positional Embedding, Identity-aware Attention, and Size Embedding on performance as a further ablation study.

Looking at the accuracy obtained on the eight ForgeryNet methods in Table \ref{tab:forgerynet_methods}, it becomes clear how both our models, and MINTIME-XC in particular, succeed in achieving excellent results on each forgery method. SlowFast R-50 appears to be particularly superior to the other models presented in recognizing video manipulated by method 8, namely ATVG-Net\cite{chen2019hierarchical}, but looking at the MAV it is evident that this result is probably achieved by simply learning the anomalies introduced by this specific forgery method. SlowFast R-50 seems to show more pronounced fluctuations in accuracy than MINTIME-XC, which remains more consistent across different methods and achieves a lower FPR.

\begin{figure}[t]
    \centering
    \includegraphics[width=0.7\linewidth]{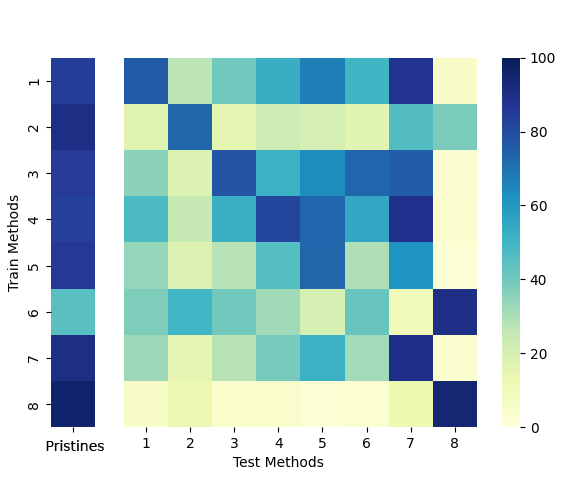}
    \caption{Accuracies obtained by MINTIME-XC on ForgeryNet for each method in cross-forgery training and testing scenario.}
    \label{fig:heatmap}
\end{figure}

\paragraph{Generalization analysis}
As pointed out in \cite{10.1145/3512732.3533582} the generalization capability of a deepfake detection model is crucial in order to be used in  the wild. In Table \ref{tab:cross_forgery}, we report the results obtained from the proposed models by training them on a subset of forgery methods and then testing them on the remaining ones. In particular, the models were trained on the so-called Identity-Remained methods, i.e. those which manipulate the video while preserving the identity of the manipulated subject, and then tested on both videos affected by Identity-Remained and by Identity-Replaced techniques (i.e. videos in which the identity was replaced). We also repeated the process but training on Identity-Replaced methods and testing on both. According to Table \ref{tab:cross_forgery}, MINTIME-XC appears to be capable of state-of-the-art results with a considerable capacity for generalization to unseen methods, which is significantly higher than that of the baseline methods. 
We conducted further analysis to see how the model behaves when trained on a single forgery method and tested on all others. For this experiment, the training set is composed by selecting all pristine videos and all the videos generated with a single forgery technique. The trained models are then tested on all the methods. The results obtained are summarized in the heatmap in Figure \ref{fig:heatmap}, where a good generalization capability is evident in most of the contexts considered. In fact, although in all contexts the model achieves higher accuracy on training methods, it still manages to recognize many videos edited with techniques not observed during training, and this is crucial in order to be able to apply it in real-world verification tasks. Method 8 differs from the others probably because of the very specific anomalies introduced in it that struggle to be recognised by our model when trained with other forgery methods.

MINTIME-XC trained on the ForgeryNet training set was then tested on the DFDC Preview test set\cite{dfdc_preview}. 
In Table \ref{tab:cross_dataset} we provide a comparison with other methods tested in this context. Although our model is trained on a different dataset from other previous works, it achieves an excellent level of generalization with high AUC score in a cross-dataset context.
\begin{table}
    \centering
    
    \resizebox{0.7\columnwidth}{!}{%
    \begin{tabular}{llll}
    \hline
        Model & Trainset & Identities & AUC \\ \hline
        Face X-ray\cite{xray} & FF++ & 1 & 65.50 \\ 
        Patch-based\cite{patch} & FF++ & 1 & 65.60 \\ 
        DSP-FWA\cite{Li_2019_CVPR_Workshops} & FF++ & 1 & 67.30 \\ 
        CSN\cite{realforensics} & FF++ & 1 & 68.10 \\ 
        Multi-Task\cite{multitask} & FF++ & 1 & 68.10 \\ 
        CNN-GRU\cite{cnngru} & FF++ & 1 & 68.90 \\ 
        Xception\cite{xception} & FF++ & 1 & 70.90 \\ 
        CNN-aug\cite{cnn-aug} & FF++ & 1 & 72.10 \\ 
        LipForensics\cite{lip} & FF++ & 1 & 73.50 \\ 
        FTCN\cite{ftcn} & FF++ & 1 & 74.00 \\ 
        RealForensics\cite{realforensics} & FF++ & 1 & 75.90 \\ \hline
        MINTIME-EF & ForgeryNet & 2 & 68.57 \\ 
        MINTIME-XC & ForgeryNet & 2 & 77.92 \\ \hline
    \end{tabular}
    }
    \caption{Cross-Dataset comparison on DFDC Preview test set. The previous methods are trained on FaceForensics++ while MINTIME-XC is trained on ForgeryNet. The identities column represents the number of considered identities during the inference.}
    \label{tab:cross_dataset}
\end{table}

\subsection{Qualitative Evaluation}
\begin{figure}[t]
\centering
\begin{subfigure}{.5\linewidth}
  \centering
  \includegraphics[width=0.9\linewidth]{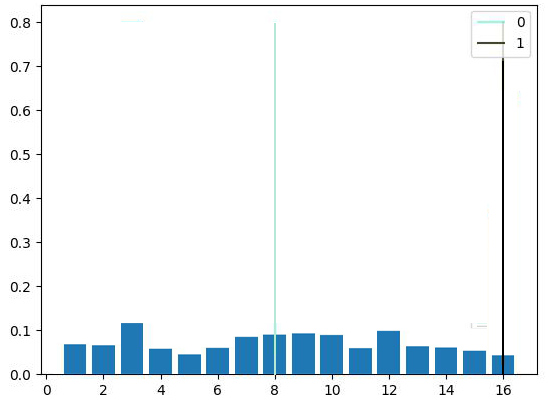}
  \caption{Pristine Video}
  \label{fig:histogram_attention_a}
\end{subfigure}%
\begin{subfigure}{.5\linewidth}
  \centering
  \includegraphics[width=0.9\linewidth]{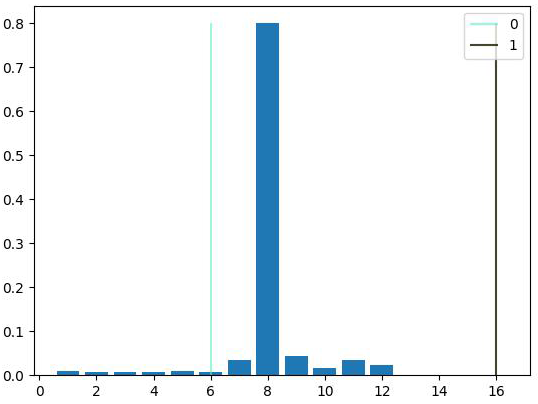}
  \caption{Fake Video}
  \label{fig:histogram_attention_b}
\end{subfigure}
\caption{Attention values histogram. On the x-axis the face position in the input sequence and on the y-axis the softmax-scaled attention values. The colored vertical lines separate the identities.}
\label{fig:histogram_attention}
\end{figure}

All our models have been trained to perform binary classification of the entire video but, in the case of deepfake videos, a hypothetical final user might be interested not only in knowing whether the video has been manipulated but also at what instant and if there is more than one tampered person. These are typical requirements when we want to expose these systems to end users (e.g. journalists) \cite{meverdeepfake}.
We can retrieve this information by analyzing the attention values obtained on the various faces which compose the input sequence. Indeed, it has been empirically shown that when the video is pristine, there are no relevant alarms of detection, as shown in Figure \ref{fig:histogram_attention_a}, while in the presence of a deepfake video, the model pays more attention on the faces containing the anomaly, as shown in Figure \ref{fig:histogram_attention_b}. By analyzing the attention values it is easy to trace which identity has been manipulated and at what instant(s) the anomaly is present.

\begin{figure}[t]
    \centering
    \begin{subfigure}{.45\linewidth}
    \includegraphics[width=0.9\linewidth]{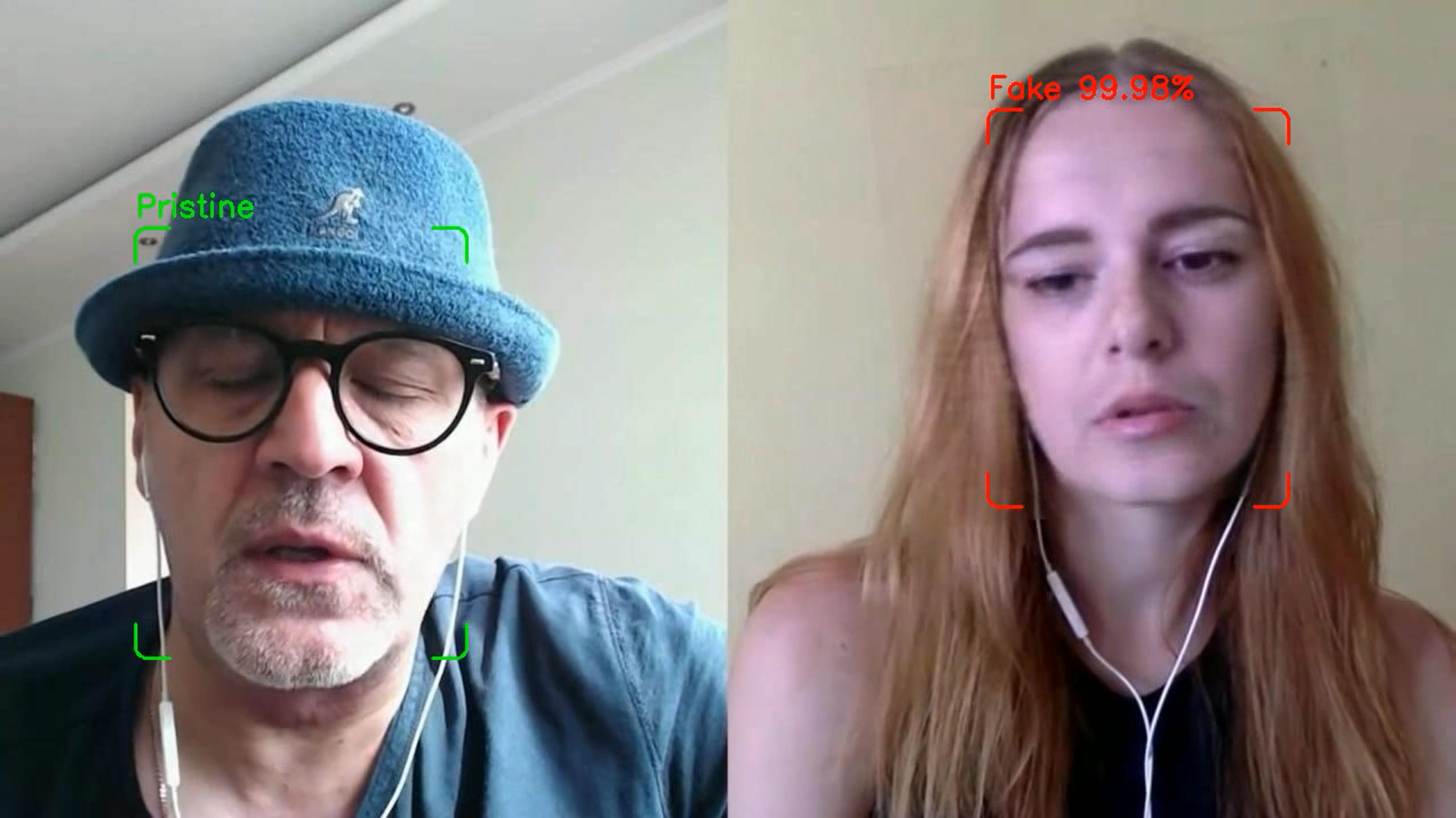}  \caption{}
    \label{fig:outcome_a}
    \end{subfigure}
    \begin{subfigure}{.45\linewidth}
    \includegraphics[width=0.9\linewidth]{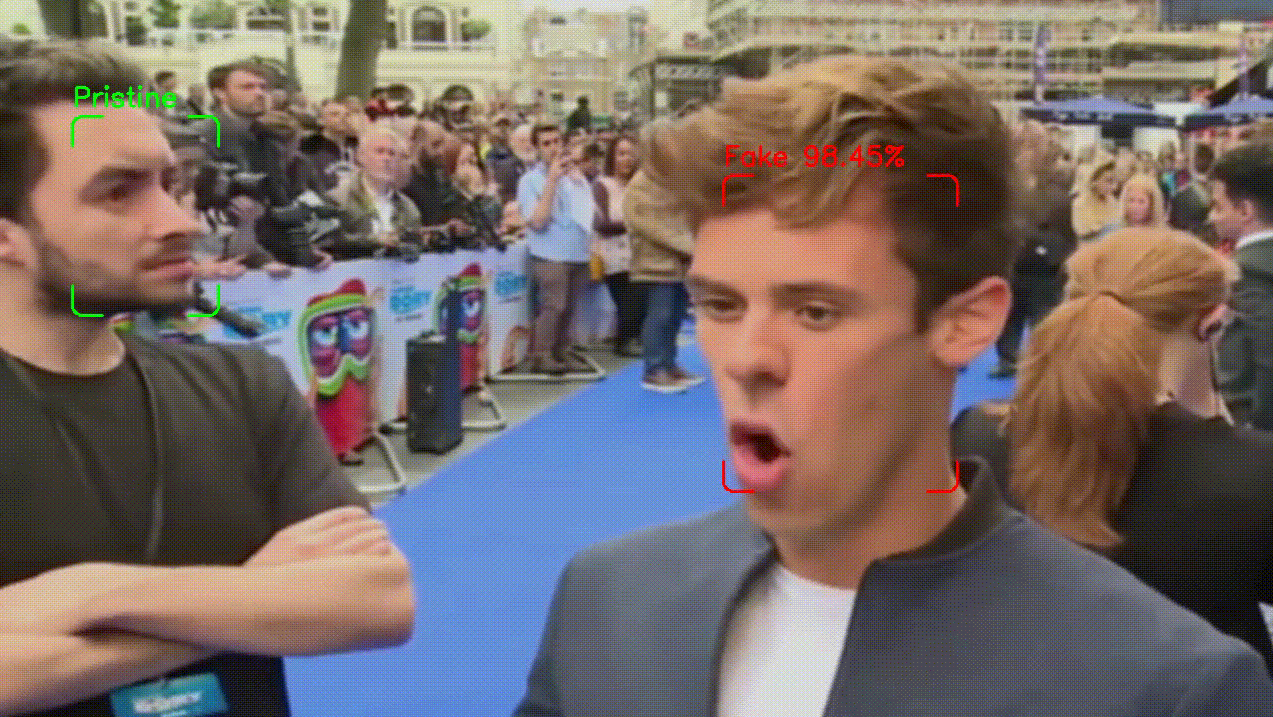}  \caption{}
    \label{fig:outcome_b}
    \end{subfigure}
    \begin{subfigure}{.45\linewidth}
    \includegraphics[width=0.9\linewidth]{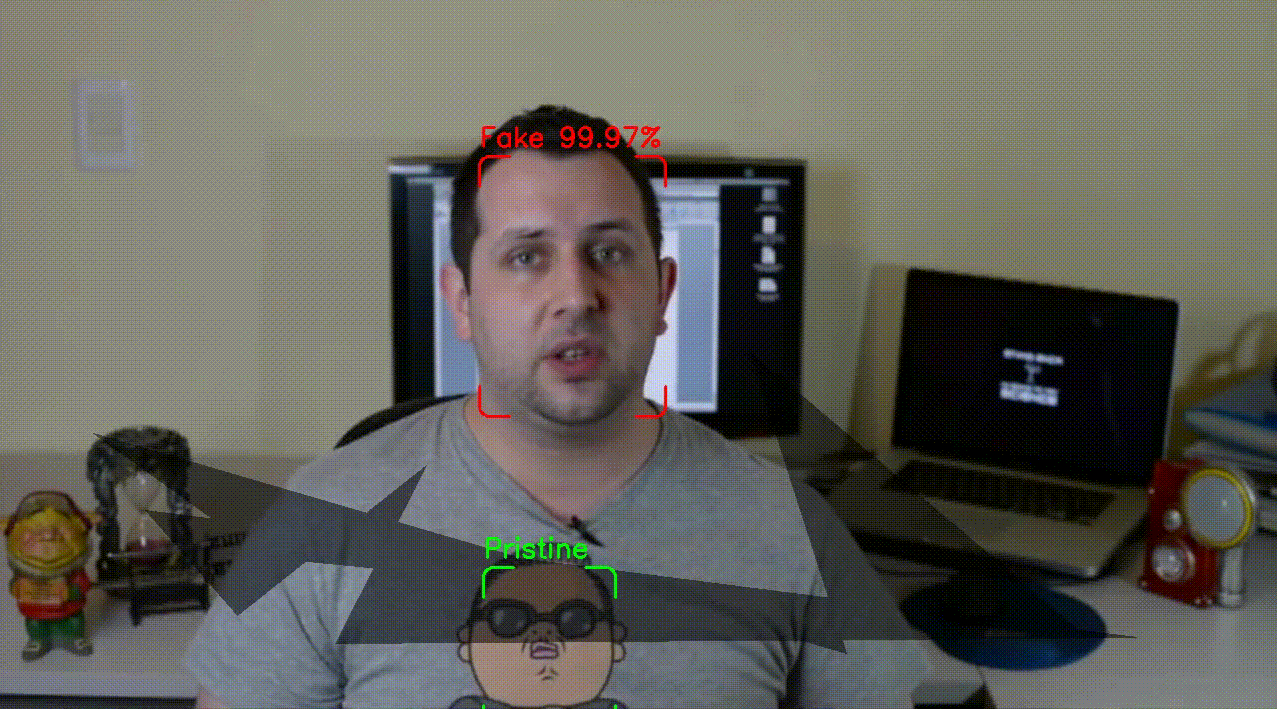}  \caption{}
    \label{fig:outcome_c}
    \end{subfigure}
    \begin{subfigure}{.45\linewidth}
    \includegraphics[width=0.9\linewidth]{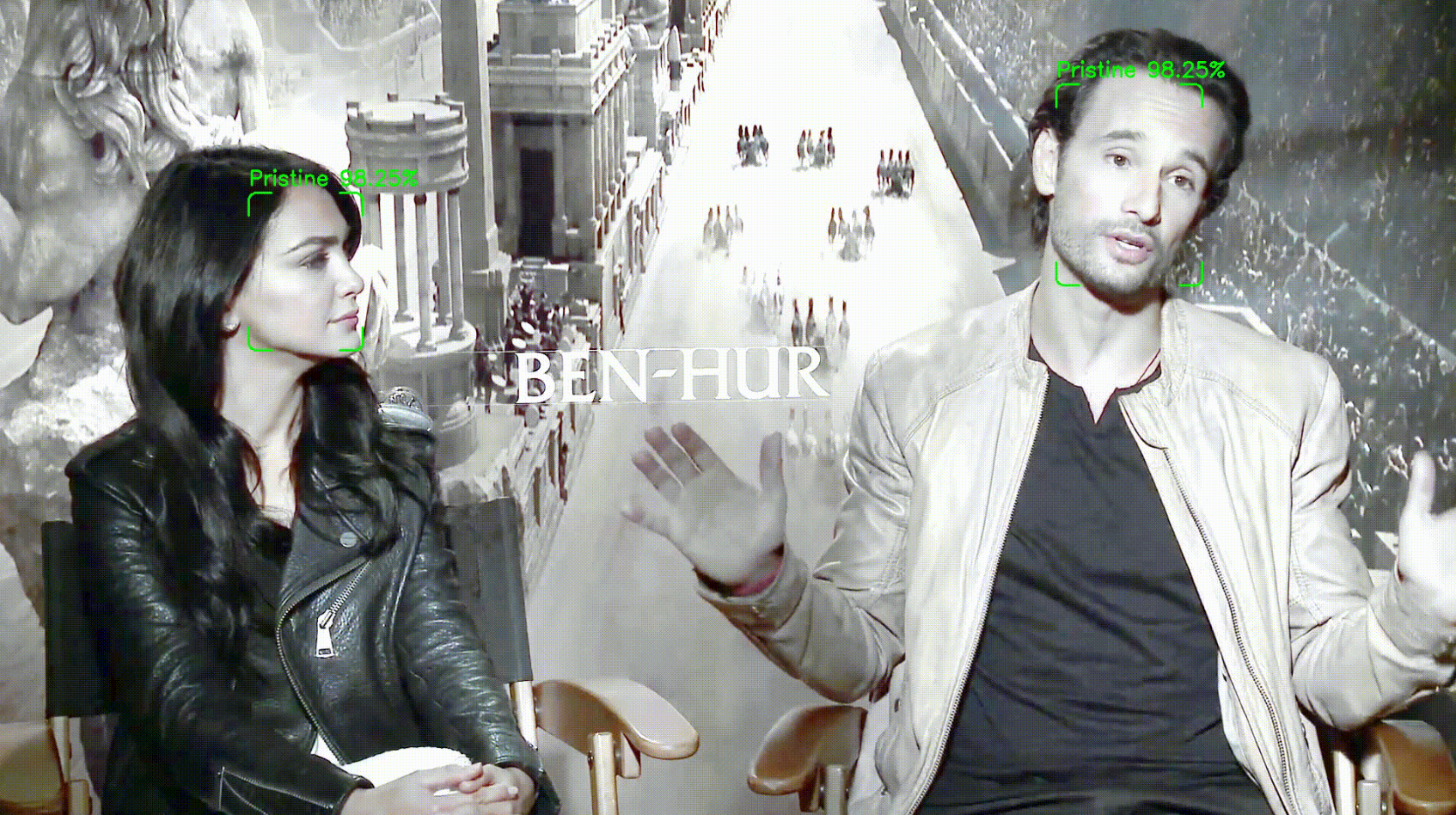}  \caption{}
    \label{fig:outcome_d}
    \end{subfigure}
    \caption{Shots of outcomes obtained with MINTIME in different multi-identity contexts.}
    \label{fig:outcome}
\end{figure}

Examples of outcomes from the model are shown in Figure \ref{fig:outcome} and it can be seen that in all cases the proposed model is able to identify the fake identity, if any, even in crowd situations like \ref{fig:outcome_b}. Interesting is the case in Figure \ref{fig:outcome_c} where there is a false face detection due to the filmed subject's t-shirt but the model still manages to realize that the manipulated face is that of the man. 

\section{Conclusions}
In this research, we identified a number of challenges encountered when trying to perform video deepfake detection in the wild and proposed an approach to overcome them. The proposed model, called MINTIME, has led to state-of-the-art results on the dataset under consideration, ForgeryNet and demonstrated high generalization level also in cross-forgery and cross-dataset contexts, often outperforming previous approaches. It can effectively handle instances of videos containing multiple people without any form of predictions aggregation, faces that vary in size relative to the total frame area, and all while capturing both spatial and temporal anomalies through the use of a modified version of TimeSformer. Finally, the trained models can be easily interpreted from the observation of attention values allowing for more fine-grained predictions and extracting useful information for the final user. As future work, it would be interesting to investigate how to leverage the audio component of videos, as a complementary input to the spatial and temporal components.

{\small
\balance
\bibliographystyle{ieee_fullname}
\bibliography{egbib}
}

\appendix
\section{Additional Preprocessing Details}
\paragraph{Face Detection and Extraction} To perform deepfake detection it is necessary to first identify and extract faces from all the videos in the dataset. Our proposed model takes as input a sequence composed of faces detected by means of a state-of-the-art face detector, i.e. MTCNN\cite{mtcnn} with an additional 30\% surface area to include a portion of the background as done in the literature\cite{10.1007/978-3-031-06433-3_19, meverdeepfake}.
To make sure that the subsequent preprocessing steps are not polluted by false detection, the face detection threshold was set at a rather high value of 95\%.
\paragraph{Identity Clustering} Having to manage multi-identity videos and wanting to detect temporal and not just spatial anomalies, it is necessary to cluster the faces in each video on the basis of their similarity and maintaining the temporal order of their appearance in the frames. To do this, a clustering algorithm was developed which groups the faces extracted from the videos into sequences. 
The algorithm is based on \cite{preprocessing} and it is structured as follows. Firstly, the features of each face are extracted via an InceptionResnetV1 pretrained on FaceNet\cite{facenet}. The similarity $s$ between each face and all the other faces identified in the video is calculated using dot product as follows:
$$s(i,j) =f_i \cdot f_j \quad \forall i, \forall j \in {0,\ldots,N-1}$$
where $f_i$ and $f_j$ are the features extracted from two distinct faces and $N$ is the total number of faces detected in a video.
A graph is constructed with hard connection if the similarity is higher than a fixed threshold and a set of clusters is constructed ignoring smallest clusters. In particular, two nodes $i$, $j$ are connected with an edge if their similarity is greater than a predefined threshold $\theta$\cite{preprocessing}. The faces inside the clusters are temporally reordered and the clusters are enumerated based on mean faces size from the largest to the smallest.
\paragraph{Data Augmentation} In order to ensure the highest level of generalization and to make the input sequences as varied as possible, a strong data augmentation was applied based on what was also applied in the creation of the ForgeryNet dataset used in our experiments. In particular, each time that a sequence is given as input to the model, 31 randomly selected transformations are applied in a uniform way for all the video, such as image compression, various types of blur techniques, image flip and invert, color editing, random noise, cutout and others. 

\section{Face-frame area ratios analysis}
\begin{figure}[t]
    \centering
    \includegraphics[width=1\linewidth]{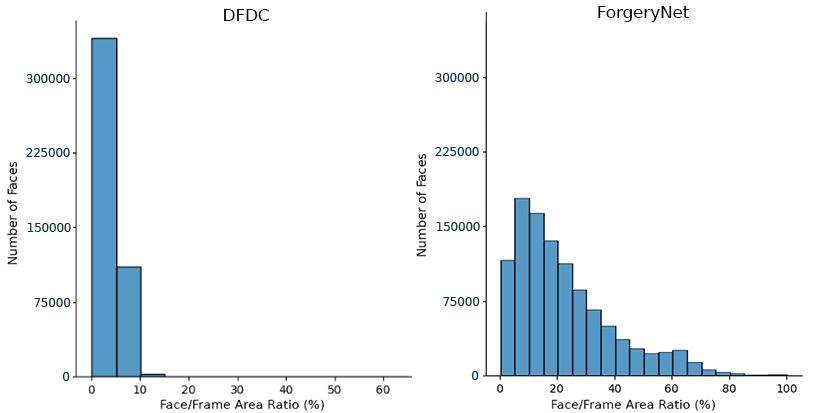}
    \caption{DFDC and ForgeryNet (training set) face-frame area ratios in detected faces.}
    \label{fig:ratios}
\end{figure}

Having to deal with the problem of varying face-frame area ratio in deepfake detection, we analysed the various available datasets to find one that contained different situations to test our model. In order to conduct this analysis, we carried out the detection of all the faces in the videos of the analyzed datasets and calculated the percentage ratio between the area occupied by the face in relation to that of the entire frame. As shown in Figure \ref{fig:ratios}, ForgeryNet\cite{forgerynet} appears to have a wide variety of face-frame area ratios clearly superior to DFDC\cite{DFDC}, which tends to have more standard shots, making it particularly suitable for our purposes.
Further statistics can be found in Table \ref{tab:datasets_statistics} where a significantly higher variance and better statistics for ForgeryNet are evident.
\begin{table}[]
    \centering
    \begin{tabular}{lllll}
    \hline
        Dataset & Min & Max & Mean & Variance \\
        \hline
         DFDC & 0.06\% & 60.1\% & 3.4\% & 5.4 \\
         ForgeryNet & 0.30\% & 100.0\% & 22.4\% & 288.6 \\
         \hline
    \end{tabular}
    \caption{Datasets statistics on face-frame area ratios of the detected faces.}
    \label{tab:datasets_statistics}
\end{table}

\section{Ablation study}
As a final analysis, to understand whether the new features introduced in the architecture are relevant to obtain the reported results, some variants of the models were trained by disabling them in part.
Size-embedding is introduced to handle certain cases that rarely occur in a deepfake detection dataset but can be very common in the real world. Nevertheless, even in a more standard context such as a dataset created specifically for deepfake detection, as Table \ref{tab:size_embedding} shows, its introduction yelds better results in terms of accuracy and AUC.
\begin{table}
    \centering
    \begin{tabular}{llll}
    \hline
        Model & Size Embedding  & Accuracy & AUC \\ \hline
        MINTIME-EF & $\times$ & 81.83 & 90.13 \\
        MINTIME-EF & \checkmark & \textbf{82.05} & \textbf{90.28} \\ 
        \hline
        MINTIME-XC & $\times$ & 87.13 & 94.03 \\
        MINTIME-XC & \checkmark & \textbf{87.64} & \textbf{94.25} \\ 
        \hline
    \end{tabular}
    \caption{Evaluation of MINTIME with and without the usage of Size Embedding on ForgeryNet Validation Set.}
    \label{tab:size_embedding}
\end{table}

Furthermore, if we consider only videos containing more than one person in the same scene, we can see in Table \ref{tab:identity_techniques} that the proposed model performs best when Multi-Identity Attention and Temporal Coherent Positional Embedding are used, demonstrating that these mechanisms contribute to a better management of multi-identity situations.

\begin{table}
    \centering
    \begin{tabular}{llllll}
    \hline
        Model & Identities & TCPE & IA & AUC & ~ \\ \hline 
        MINTIME-XC & 2 & $\times$ & $\times$ & 93.29 & ~ \\ 
        MINTIME-XC & 2 & \checkmark & \checkmark & \textbf{94.12} & ~ \\ \hline
        MINTIME-XC & 3 & $\times$ & $\times$ & 90.57 & ~ \\ 
        MINTIME-XC & 3 & \checkmark & \checkmark & \textbf{93.32} & ~ \\ \hline
    \end{tabular}
    \caption{Evaluation of MINTIME-XC with and without the usage of identity-aware techniques on multi-identity videos only from ForgeryNet Validation Set. The Identities column indicate the number of identities considered during inference.}
    \label{tab:identity_techniques}
\end{table}

Finally, in order to explore the impact of the identity reordering policy on the results and ensure that no bias was induced during the training phase, we carried out some tests using three different techniques. When constructing the input sequence, it is in fact necessary to decide in which order to insert the various identities, in case of more than one, inside it. In the training phase, the average of the areas of the faces associated with each identity was used as a criterion (Size-Based), giving greater importance to the most prominent identities in the scene. Therefore, an identity containing more prominent faces in the video would have been inserted earlier in the sequence and would potentially have occupied a larger portion of it. As a test, at inference time, we also tried using the number of faces associated with an identity as a criterion, thus giving more weight to the most frequent identities (Frequence-Based) as done in \cite{forgerynet}, and finally we also compared a random criterion.
As can be seen from Table \ref{tab:ordering}, the three different strategies do not particularly affect performance of our proposed model and the only one that has a slightly significant loss is Random-Based, probably due to the fact that in some rare cases, by reordering the identities randomly and selecting the top two, the fake identity could be discarded in favour of other identities in the scene. Quite the opposite situation on a SlowFast, which, considering a single identity per video is strongly affected by the choice of the latter. The results are therefore very different from case to case, leading to better results in random-ordering simply by chance in the choice of the fake identity.

\begin{table}
    \centering
    \begin{tabular}{llll}
    \hline
        Model & Sorting Method & Accuracy & AUC \\ \hline
        SlowFast R-50 & Random & 77.02 & 85.97 \\
        SlowFast R-50 & Frequence-Based & 76.04 & 84.58 \\
        SlowFast R-50 & Size-Based & 72.63 & 80.92 \\
        MINTIME-XC & Random & 86.25 & 93.73 \\
        MINTIME-XC & Frequence-Based & 86.43 & 94.08 \\
        MINTIME-XC & Size-Based & \textbf{86.68} & \textbf{94.12} \\ \hline
    \end{tabular}
    \caption{The impact of identities sorting techniques at inference time on multi-identity videos only from ForgeryNet Validation Set. All the methods are trained with Size-based approach.}
    \label{tab:ordering}
\end{table}

\end{document}